\documentclass[runningheads]{llncs}
\usepackage{graphicx}

\usepackage{tikz}
\usepackage{comment}
\usepackage{amsmath,amssymb} 
\usepackage{color}

\usepackage[accsupp]{axessibility}  

\usepackage{amsmath}
\usepackage{amssymb}
\usepackage{booktabs}
\usepackage{multirow}
\usepackage{bm}
\usepackage[breaklinks=true,
            bookmarks=false,
            colorlinks,
            linkcolor=red,       
            anchorcolor=blue,  
            citecolor=green, 
            ]{hyperref}
\usepackage[shortcuts]{extdash}

\usepackage[capitalize]{cleveref}
\crefname{section}{Sec.}{Secs.}
\Crefname{section}{Section}{Sections}
\Crefname{table}{Table}{Tables}
\crefname{table}{Tab.}{Tabs.}

\renewcommand{\paragraph}[1]{\noindent{\bf #1}}
\renewcommand{\subparagraph}[1]{\noindent{\it #1}}

\usepackage{amssymb}
\usepackage{pifont}
\newcommand{\cmark}{\ding{51}}%
\newcommand{\xmark}{\ding{55}}%

\newcommand{\sslshortname}{{SOS}}
\newcommand{\netname}{{OIC}}

\newcommand{\stageonetrainname}{Stage I}
\newcommand{\stagetwotrainname}{Stage II}

\newcommand{\eg}{{\it e.g.}}
\newcommand{\ie}{{\it i.e.}}
\newcommand{\cf}{{\it cf.}}
\newcommand{\wrt}{w.r.t}

\begin{document}
\pagestyle{headings}
\mainmatter
\def\ECCVSubNumber{824}  

\title{SOS! Self-supervised Learning Over Sets Of Handled Objects In Egocentric Action Recognition} 


\titlerunning{SOS: Self-supervised Learning Over Sets}
%
\author{Victor Escorcia\inst{1}
\and
Ricardo Guerrero\inst{1}
\and
Xiatian Zhu\inst{2}
\and
Brais Martinez\inst{1}
}

\authorrunning{Escorcia et al.}
%
\institute{Samsung AI Center Cambridge$^1$, University of Surrey$^2$\\
\email{\{v.castillo, r.guerrero, brais.a\}@samsung.com}}
\maketitle

\begin{abstract}
Learning an egocentric action recognition model from video data is challenging due to distractors in the background, \eg, irrelevant objects. Further integrating object information into an action model is hence beneficial. Existing methods often leverage a generic object detector to identify and represent the objects in the scene. However, several important issues remain. Object class annotations of good quality for the target domain (dataset) are still required for learning good object representation. 
Moreover, previous methods deeply couple existing action models with object representations, and thus need to \mbox{retrain} them jointly, leading to costly and inflexible integration. 
To overcome both limitations, we introduce \mbox{{\bf S}elf-Supervised} Learning {\bf O}ver {\bf S}ets (\sslshortname),
an approach to \mbox{pre-train} a generic Objects In Contact (\netname) representation model from video object regions detected by an \mbox{off-the-shelf} hand-object contact detector.
Instead of augmenting object regions individually as in conventional \mbox{self-supervised} learning, we view the action process as a means of natural data transformations with unique \mbox{spatio-temporal} continuity and exploit the inherent relationships among per-video object sets. Extensive experiments on two datasets, EPIC-KITCHENS-100 and EGTEA, show that our \netname{} significantly boosts the performance of multiple \mbox{state-of-the-art} video classification models.

\end{abstract}

\section{Introduction}
\label{sec:intro}

Egocentric videos recorded by wearable cameras give a unique perspective on human behaviors and the scene context \cite{damen2018scaling,li2018eye,sigurdsson2018actor}.
Existing action recognition methods, typically developed for third-person video understanding, focus on learning from the whole video frames~\cite{carreira2017quo,zhou2018temporal,lin2019temporal,wang2018temporal,bulat2021space,feichtenhofer2019slowfast,kazakos2019epic}.
Unlike third-person videos where actions often take place in dramatically different background scenes (e.g., swimming in a pool, cycling on the road) \cite{carreira2017quo,soomro2012ucf101}, egocentric videos are often collected at a specific scene (e.g., a kitchen) with similar background shared across different human actions (e.g., cutting onion and washing knife) and cluttered with distractors (e.g., a knife on the countertop).
These distinctive characteristics bring about extra challenges for most existing action models {\em without a fine-grained understanding of spatio-temporal dynamics and context.}

\begin{figure}
    \centering
    \includegraphics[width=\linewidth]{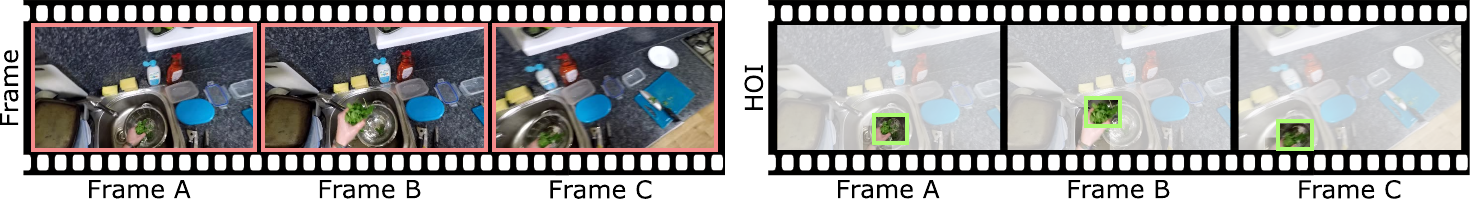}
    \vskip -0.3cm
    \caption{{\bf Handled Objects are vital players for determining actions in egocentric videos.} However, most action classification models learn directly from video frames (left). This paper puts manipulated objects (right) 
    at the forefront {\em without} the need for expensive and tedious fine-grained object annotations.
    }
    \label{fig:example_hoi}
    \vspace{-0.5cm}
\end{figure}


To overcome these challenges, recent works have started to incorporate information from other modalities, such as audio \cite{kazakos2019epic,kazakos2021slow}, narration language \cite{kazakos2021little,bertasius2020cobe}, and eye gaze \cite{li2021eye}, into the video action model.
This is motivated by their complementary information 
and
the modality-specific nature of distractors. 
Object information was also shown to be a powerful complement to prior video action models \cite{wu2019long,wang2020symbiotic}. While a video model represents the sequence as a whole, combining foreground and context, object detectors exploit bounding box annotations to explicitly model each object, separately from the
others
and background. 

Our work falls in this line of research with a crucial difference. We consider that bounding box annotations are costly and unsuited for {\em long-tail open-vocabulary object distributions} across the scenes, e.g., a kitchen setting. Large-scale annotations are thus seldom possible in most cases. We tackle this issue by capitalizing {\em an off-the-shelf hand-object contact detector}
to localize class-agnostic object regions in video.
Crucially, we introduce a novel self-supervised approach to learn a specialized representation on detected object regions without resorting to object box annotations nor labels.
Our key idea is to leverage the spatio-temporal continuity in videos as {\em a native context constraint} for exploiting the inherent relationships among the set of detected object regions per video. Intuitively, all the handled object regions play respective roles during the course of an action, and they collectively provide a potentially useful context clue for learning a suitable object representation (e.g., the ``tap" and ``plate" while washing dishes - see Fig. \ref{fig:qualitative}). 
With our class agnostic self-supervised learning, there is also a potential that the natural action class imbalance problem would be simultaneously alleviated -- a typical yet understudied problem in video understanding.
We term our self-supervised pre-training {\bf S}elf-Supervised Learning {\bf O}ver {\bf S}ets
(\sslshortname). 
After pre-training, we transfer the specialized representation of detected objects to a target task, e.g., video classification. For this purpose, we employ an Objects in Contact (\netname) network and further fine-tune the entire representation with {\em weak} video-level labels.
Once trained, our \netname{} can be flexibly integrated with existing video classification models, further boosting their performance.

We make three {\bf contributions} in this paper.
{\bf(1)} We investigate the merits on learning a representation of handled objects for egocentric action recognition,
from the perspective of boosting state-of-the-art video action models in a flexible manner without the need for expensive fine-grained bounding-box labeling on the target dataset.
{\bf(2)} To that end, we leverage an off-the-shelf hand-object contact detector to generate class-agnostic object regions and introduce a novel set self-supervised learning approach, \sslshortname, to learn a specialized representation of object regions. \sslshortname{} exploits the inherent relation (e.g., concurrence) among handled objects per video to mine the underlying action context clue by treating all the objects collectively as a single set.
{\bf(3)}
Experiments on two egocentric video datasets showcase that our \netname, aided with our \sslshortname{} pre-training, complements multiple existing video classification network and yields state-of-the-art results in video classification. Moreover, we demonstrate the benefit of \sslshortname{} for dealing with the realistic long-tail setup in videos. (Sec. \ref{exp:sos_lt}).

\section{Related Work}

\noindent{\bf Egocentric action recognition. }
Egocentric action recognition has made significant advances in recent years, thanks to the introduction of ever larger video benchmarks \cite{damen2018scaling,damen2020rescaling,li2018eye}.
Early efforts were focused on adapting representative generic video models \cite{wang2018temporal,lin2019temporal,zhou2018temporal,feichtenhofer2019slowfast}.
Later on, a variety of dimensions have been investigated.
For example,
Kazakos et al. \cite{kazakos2019epic} combined multi-modal information (e.g., optical flow, audio) within a range of temporal offsets.
Bertasius et al. \cite{bertasius2020cobe} leveraged
the language based semantic context to supervise
the learning of active object detection.
Li et al. \cite{li2021ego} investigated the pre-training of video encoder for mitigating the domain gap from 
the common pre-trained video datasets (e.g., Kinetics \cite{carreira2017quo}). 
Sudhakaran et al. \cite{sudhakaran2019lsta}
designed an LSTM based Long-Short Term Attention model to 
locate relevant spatial parts (e.g., active objects)
with temporally smooth attention.
Similarly, Yan et al. \cite{yan2020interactive}
proposed an Interactive Prototype Learning (IPL) model for
better learning active object representations by interaction with different motion patterns.
Instead, Li et al. \cite{li2018eye,li2021eye} and Liu et al. \cite{liu2020forecasting} used human gaze to guide the attention of deep models to interacting regions.
%
Similar to ours, object detection has been 
previously exploited for improving action
recognition. Indeed, early work already identified explicit hand detection as an informative queue for action recognition \cite{lendingahand2015iccv}. More recently, Wang et al. \cite{wang2018videos} exploited object regions and their spatio-temporal relations to enhance video representation learning. More similar to ours, Baradel et al. \cite{Baradel_2018_ECCV} devised an architecture with a video branch and an object branch. However, their approach requires object-level annotations and has no mechanism to identify foreground/active objects, thus being vulnerable to distractors. Wu et al. \cite{wu2019long} instead used an attentive mechanism and also incorporated long-term temporal context. Wang et al. \cite{wang2020symbiotic} proposed a model specific for egocentric action recognition that relies on a complex attentive mechanism to sieve out distractors. 
Unlike prior work, our method does not assume fine-grained object-level annotations from the target dataset, hence being scalable in practical applications.
Critically, their object representation is tightly coupled with an action model
with the need for joint training.
In contrast, we learn an independent object representation model
enabling to flexibly benefit off-the-shelf action models in a decoupled post-training manner.


\paragraph{Hand-object interaction (HOI).} While generic Human-Object Interaction is a widely-studied topic \cite{gkioxari2018detecting,qi2018learning,li2020hoi,hou2020visual}, recent works have shown that it is possible to train large-scale domain-agnostic hand and hand-object contact detectors \cite{shan2020understanding,narasimhaswamy2020detecting}. As face detection models, these models can be deployed without domain-specific retraining and still maintain reasonable effectiveness. In our work, we apply one such off-the-shelf hand-object contact model without retraining \cite{shan2020understanding}.









\noindent{\bf Self-supervised learning. }
There is a recent surge in self-supervised learning (SSL)
for learning generic feature representation models
from large scale dataset without labels \cite{he2019momentum,chen2020exploring,chen2020simple,caron2020unsupervised,grill2020bootstrap,SunNTS2021}.
We are inspired by this trend and introduce a
novel application of exiting SSL techniques that learns a representation specific of objects being manipulated.
However, a direct application of a SSL model to our problem does not produce optimal results. We propose two important modifications:
firstly, instead of following a classic fine-tuning from domain-agnostic pre-training, e.g. over ImageNet,
our method leverages on-domain SSL for solving the domain shift problem with model pre-training.
To this end we use a domain-agnostic SSL model as means of pre-training, followed by a domain-specific SSL training \cite{ssl4ssl}. 
Secondly, existing SSL methods focus on single images, using two different augmentations to obtain two copies of an image. Instead, in the presence of video, sampling different timestamps and locations can create more natural and effective augmentations \cite{demystifyingneurips20}. Inspired by this insight, we propose a variant of SwAV that operates on sets of image regions extracted from a video sequence. The experimental results section shows that these two modifications are crucial to obtain an effective object representation.

\noindent{\bf Long-tail learning in video.}
Real-world egocentric actions 
are typically class-imbalanced \cite{damen2018scaling,damen2020rescaling,li2018eye}.
Despite extensive works in image domains
\cite{dong2018imbalanced,huang2016learning,kang2020exploring,kang2019decoupling,menon2020long,yang2020rethinking},
class imbalance is still less studied in video tasks \cite{zhang2021videolt}.
Inspired by the intriguing finding that unlabeled image data is effective in counting class imbalance via SSL  \cite{yang2020rethinking}, this work presents an object set based SSL method. 
Unlike \cite{yang2020rethinking} focusing on images,
we consider more complex video data due to
the extra temporal dimension, the structured nature of action labels (\ie, defined as tuple of two imbalanced labels distributions, verb and noun) and only access to video-level supervision.
This paper contributes the first study of the relationship between SSL and long-tailed learning in egocentric video understanding.
%


\section{Method}

\begin{figure}[!t]
    \centering
    \includegraphics[width=0.75\linewidth]{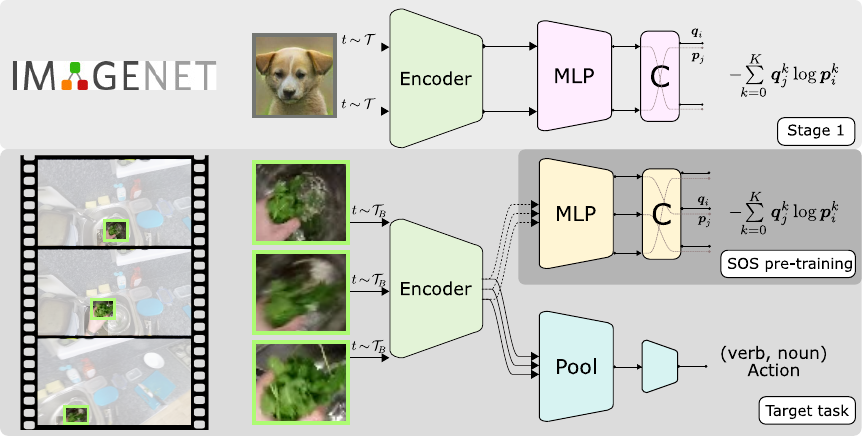}
    \vskip -0.3cm
    \caption{
    {\bf Overview of the proposed Self-Supervised Learning Over Sets (\sslshortname) approach }
    for pre-training a specialized Objects In Contact (\netname) representation model. 
    Taking as input object regions extracted by an off-the-shelf hand-object contact detector in videos, we formulate a two-staged pre-training strategy.
    The {\em first} stage consists of generic self-supervision learning (e.g. ImageNet).
    In the {\em second} stage, we exploit self-supervised learning on per-video object sets using actions as the natural transformations
    with spatio-temporal continuity.
   Given a target task, we further fine-tune the representation through {\em video-level} action labels.
    }
    \label{fig:pipeline}
    \vspace{-0.5cm}
\end{figure}

We aim to learn a generic object-in-contact (\netname{}) model for improving the performance of existing egocentric action models in a plug-and-play manner.

\paragraph{Overview.}
Given a video $V$ and an off-the-shelf hand-object contact detector model \cite{shan2020understanding}, 
we obtain a set of object regions that likely contain objects manipulated by the hands.
Let $\mathcal{B}=\{\bm{B}_i\}_{i=1:M}$ denote the set of $M$ object regions in the video.
We have an object region encoder $f_B$ so that $\bm{y}_B^i=f_{B}(V,\bm{B}_i;\theta_B)$.
To that end, we propose a simple yet effective approach termed, {\bf S}elf-Supervised Learning {\bf O}ver {\bf S}ets (\sslshortname), to train $\theta_B$ in two steps: 
{\bf (I)} First, a large-scale self-supervised learning model is used for pre-training. 
{\bf (II)} Followed by, an on-domain self-supervised learning stage yielding a specialized representation better suited for the task of interest. 
Given a target task, standard discriminative fine-tuning is then followed, using the video-level labels and standard cross-entropy as supervision.
An overview of our \sslshortname{} is depicted in Fig. \ref{fig:pipeline}.

\subsection{Self-Supervised Object Representation Learning from Video Object Regions}
\label{sec:ssl}

Object representations are typically learned as part of the standard object detector training pipeline. However, we do not assume the availability of object-level annotations. We propose to instead learn the object representation in an unsupervised manner using object regions.

\begin{figure}[!t]
    \centering
    \includegraphics[width=\linewidth]{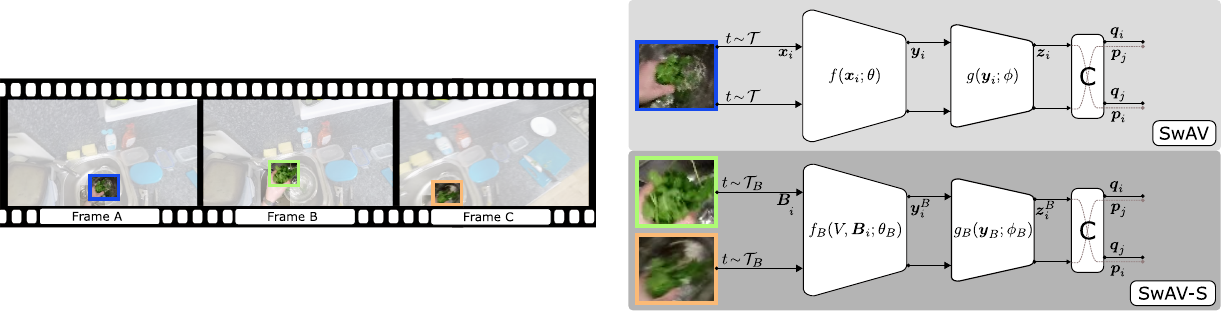}
    \vskip -0.3cm
    \caption{
    {\bf Overview of the proposed SwAV-S.} Schematically outlining the differences \wrt the standard SwAV~\cite{caron2020unsupervised} for image regions, and our self-supervised pre-training approach using videos and leveraging actions as natural transformations.
    }
    \label{fig:swavs}
    \vspace{-0.5cm}
\end{figure}

\paragraph{\stageonetrainname{}: Model pre-training.}
There is a lack of standard large-scale datasets for model pre-training in the egocentric video, so one could use the widely used
ImageNet~\cite{russakovsky2015imagenet} supervised pre-trained weights for model initialization.
This gives rise to an inevitable domain shift challenge
for object representation learning due to the intrinsic discrepancy
in data distribution.
To overcome these domain shift challenges,
we leverage the more domain-generic self-supervised learning (SSL) strategy
for model pre-training \cite{chen2020simple,grill2020bootstrap,he2019momentum,vondrick2018tracking,chen2020exploring,caron2020unsupervised}.
In practice, due to the relatively small size of the target dataset, it is key to start with an SSL model trained on ImageNet for model initialization. 
In general, any existing SSL method is applicable. Based on preliminary experiments, we select the recent state-of-the-art model SwAV \cite{caron2020unsupervised} (Fig. \ref{fig:swavs} top - \stageonetrainname{}). 



\paragraph{Stage II: On-domain Self-supervised learning Over Sets, \sslshortname{}.} 
The key challenge is how to capitalize the unlabeled object regions. 
Motivated by the results of \cite{ssl4ssl}, we perform on-domain SSL to specialize the object region encoder $f_B$ to better represent the regions in the target domain, egocentric videos.

It is worth noting that the promising recipe from~\cite{ssl4ssl} ignores a key and important piece of information from our problem definition, object regions extracted from a single video are {\em not} independent but 
correlate to an ongoing action instance {\em collectively}.
Inspired by this consideration, we introduce the notion of actions as natural data transformation exploiting the unique spatio-temporal continuity of videos and the inherent relationships among per-video object sets.
Intuitively, the set of object regions captures the underlying action context in the video. We argue that statistically predominant inter-object relationships (\eg, concurrence) provide a strong self-supervisory signal for learning an specialized object representation.



We consider each video $V$ as a training sample and treat the {\em set construction process from all regions in the video $\mathcal{B}$} as {\em a spatio-temporal transformation process}, subject to the structural variations of the performed action in the video (\ie, elastic, geometrical or ambient changes that objects undergo through space and time).
%
%
This is conceptually reminiscent to and complement the standard image augmentation process (e.g., cropping, flipping, jittering).
Interestingly, under this perspective, our approach fits exactly on
recent top-of-the-line self-supervised frameworks such as contrasting learning, clustering, instance similarity, decorrelated representation, to name a few \cite{chen2020simple,grill2020bootstrap,he2019momentum,chen2020exploring,caron2020unsupervised}.
%
%
Without loss of generality, we present our Self-supervised Learning Over Sets approach (\sslshortname) on top of the SwAV formulation due to their state-of-the-art results. Yet, we anticipate that our \sslshortname\ could be extended to other frameworks\cite{chen2020simple,grill2020bootstrap,he2019momentum,chen2020exploring}.


\noindent{\bf SwAV review.} 
SwAV \cite{caron2020unsupervised} aims at learning a feature representation that matches different views of the same image.
Specifically, the representation of one of these views is used to compute a code $\bm{q}$, which is then predicted from the code of another view. Formally, given an image $\bm{x}$, two corresponding views $\bm{x}_1$ and $\bm{x}_2$ are created by applying a random transformation $t\in \mathcal{T}$, where $\mathcal{T}$ is the set of all considered transformations (typically these are synthetic, such as random crops, color jittering, Gaussian blurring and flipping,
etc.).
%
Corresponding feature vectors $\bm{z}_i=g(f(\bm{x}_i;\theta);\phi)$ are generated and projected into the unit sphere, where $\bm{z}_i \in \mathbb{R}^d$, $f$ represents a backbone network with parameters $\theta$ and $g$ a projection head with parameters $\phi$. SwAV can be seen as an online clustering method, where the cluster centroids are defined by set of learnable prototypes 
$\bm{c}_i \in \bm{C}^{K \textup{x} d}$, which are used as a linear mapping function to compute each view's code $\bm{q}_i = \bm{z}_{i}^{\top}\bm{C}$. 
The objective is to predict a view's code $\bm{q}_i$ from the other view's features $\bm{z}_j$, and the problem is formulated via cross-entropy minimization as:
\begin{equation}
\label{eq:swav_loss}
    L(\bm{z}_1,\bm{z}_2) = \ell(\bm{q}_1,\bm{z}_2) + \ell(\bm{q}_2,\bm{z}_1)
\end{equation}

\noindent where the first term is
\begin{equation}
    \ell(\bm{q}_1,\bm{z}_2) = - \sum_{k=0}^{K} \bm{q}_1^k \, \textup{log} \, \bm{p}_2^k \;,
\end{equation}
where $\bm{q}_1$ represents the prototype likelihood or soft cluster assignment of $\bm{x}_1$, $\bm{p}_2^k$ represents the prediction of $\bm{q}_1$ as the softmax of
$\bm{z}_{2}^{\top}\bm{c}_{k} / \tau$ and $\tau$ is a temperature parameter. Additionally, $\bm{q}_1^k$ comes from $\bm{Q}$, which normalizes all $\bm{q}$ in a mini-batch (or queue) using the Sinkhorn-Knopp algorithm. The second term of Eq. \eqref{eq:swav_loss} is similarly defined.
In practice, SwAV uses more than two views per image. More specifically, the concept of multi-crop is introduced, where crops taken as part of the augmentation strategy can be either global or local concepts of an image. Furthermore, the problem is formulated in such a way that prototype code $\bm{q}$ assignments are only done on global views, while predictions $\bm{p}$ are done using both local and global views.

\noindent {\bf Our SwAV-S.}
To better exploit unlabeled object regions from videos, we introduce a set structure into SwAV's formulation, resulting in a new SSL variant 
dubbed as {\em SwAV-S}.
%
%
Specifically, we sample a subset of object regions from $\mathcal{B}^{'} =  \{\bm{B}_i\}_{i=1:N}, \mathcal{B}^{'} \subset \mathcal{B}$ from each video. Then, each region undergoes an independent image transformation and is treated as one view of $V$ to be predicted 
(or contrasted) 
from another region of the same set. This set of regions are encoded, generating embedding vectors $\{\bm{z}_i=g_B(f_B(V,\bm{B}_i; \theta_B);\phi_B)\}_{i=1:N}$, where $f_B$ and $g_B$ represent a non-linear encoder function and projection head, respectively. 

In contrastive learning design, we make a couple of important differences against original SwAV. 
(1) At each training iteration, we sample $N>2$ object regions $\{B_i\}_{i=1:N}$ from a video sequence $V$. 
(2) We only consider global views, which allows the expansion of terms in Eq. \eqref{eq:swav_loss} to $N$, effectively treating all object regions as a set as follows:
\begin{equation}
\label{eq:swavs_loss}
    L = \frac{-1\hphantom{-}}{N^2\!-\!N} \sum^N_i \sum^{N}_{j \neq i}  \sum_{k=0}^{K} \bm{q}_i^k \, \textup{log} \!
    \left ( \! \frac {\textup{exp} \left ( \frac{1}{\tau}\bm{z}_{j}^{\top}\bm{c}_{k}  \right )}{\sum_{k'}\textup{exp} \left ( \frac{1}{\tau}\bm{z}_{j}^{\top}\bm{c}_{k'}  \right )} \! \right )
\end{equation}
where ($i,j$) indexes the pairs of regions from a video.



\paragraph{Discussion.} While our approach shares some high-level ideas \wrt\ earlier SSL algorithms \cite{wangHG2017}, \sslshortname{} relies on (1) different assumptions (\eg, do not require an explicit graph of relations among patches or tracking), (2) different aim (integrating object representations for action recognition), and (3) specific methodology (self-supervised learning over sets). All in all, our approach revives this line of research with a refreshing perspective and using leveraging recent insights.

\subsection{Target task finetuning: \netname{} Net and Model Fusion}
\label{sec:aggr}
\begin{figure}[t]
    \centering
    \includegraphics[width=\linewidth]{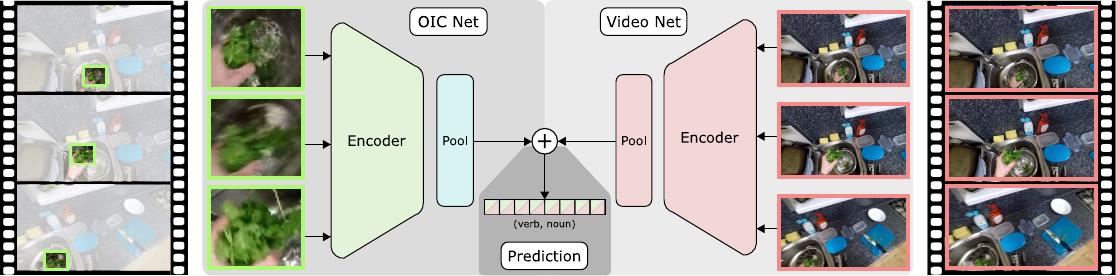}
    \vskip -0.3cm
    \caption{
    {\bf Our \netname{} model can complement any existing video classification network, Video Net.}
    While most 
    existing classification networks consume the entire video frames, ours \netname{} only takes object regions. 
    Their predictions are then late fused.
    }
    \label{fig:fuse}
    \vspace{-0.5cm}
\end{figure}

Following the typical transfer learning of SSL pipelines, The pre-trained enhanced object region encoder $\theta_B$ serves as initialization for the \netname{} encoder $f_B$ addressing the target task (\cf\ Fig. \ref{fig:fuse} - \netname{} Net). In our case, the target task corresponds to supervised video classification by means of object patches.



Given a set of object regions $\mathcal{B}$ per video,
we feed them through the \netname{} encoder, followed by a pooling module aggregating the information from all the cohort to obtain the classification logits of interest $l= h(f_{B}(V,\mathcal{B};\theta)) \in \mathbb{R}^{\lvert\mathcal{Y}\rvert}$, where $\lvert\mathcal{Y}\rvert$ corresponds to the size of the label space.
Note that we only have access to the weak video-level supervision. We apply the standard cross-entropy loss between the logits and the video-level label to optimize the network parameters.

\paragraph{Class imbalanced finetuning.}
To complement our \sslshortname{} pre-training for tackling the class imbalance issue,
we adopt the recent logit adjustment method \cite{menon2020long} with
the key idea to impose the class distribution prior of training data through logit regulation.
For training, we optimize a logit adjustment cross-entropy loss, 
$L_{la} = -\log\frac{e^{f_y + \tau\log{\pi_y}}}
{\sum_{y' \in \mathcal{Y}} e^{f_{y'} + \tau\log{\pi_{y'}}}}$
with $\pi_y$ the frequency of class $y$ on the training set (i.e., the class prior),
and $\tau$ the temperature.
In our case, we impose this adjustment to the noun and verb branches separately, so that their imbalance can be remedied according to their respective distribution priors.

\paragraph{Model Fusion.}
Once trained as discussed above, 
our \netname{} model can be integrated with 
any existing action models \cite{bulat2021space,lin2019temporal,feichtenhofer2019slowfast}
as depicted in Fig. \ref{fig:fuse}.
For simplicity and flexibility, we use a weighted late-fusion of video predictions as:
\begin{equation}
\bm{l} = \alpha_{\text{\netname{}}} \bm{l}_{\text{\netname{}}} + \alpha_{i} \bm{l}_i,
\label{eq:fuse}
\end{equation}
where $\bm{l}_{\text{OCI}}, \bm{l}_i$ are the classification logits of our \netname{} and any existing model; and $\alpha_{\text{OCI}}, \alpha_i$ are scalars weighting the confidence of each model. In practice, we set $\alpha_{\text{OCI}}, \alpha_i$ based on the performance of each individual model on a pilot validation set (\eg, 30\% of training set). In the case of verb and noun based action prediction, the fusion is applied separately.

\section{Experiments}

\paragraph{Datasets.} 
We evaluate our method on two standard egocentric video datasets in the domain of kitchen environments.
{\bf \em EPIC-KITCHENS-100} \cite{damen2020rescaling} is a large-scale egocentric action recognition dataset with more than 90,000 action video clips extracted from 100 hours of videos in kitchen environments.
It captures a wide variety of daily activities in kitchens. 
The dataset is labeled using 97 verb and 300 noun classes
and their combinations define different action classes. 
Please refer to the \textbf{supplementary material} for results in the test server, additional details, among others.
{\bf \em EGTEA} \cite{li2018eye} is another popular egocentric video dataset consisting of 10,321 video clips annotated with 19 verb classes, 51 noun classes, and 106 action classes. 

\paragraph{Performance metrics.}
We report 
the 
model evaluation results using the standard action recognition protocol \cite{damen2020rescaling,li2018eye}.
Specifically, each model predicts 
the 
verb and 
the 
noun using two heads,
and the accuracy rates of verb, noun, and action (\ie\ verb and noun tuple) are calculated as performance metric.
For EPIC-KITCHENS-100, we report the standard accuracy on different set of action classes.
For EGTEA, we adopt the mean class accuracy on the three train/test splits.

\paragraph{Implementation details.}
We use an off-the-shelf hand-object contact detector to extract candidate regions of interest where it is likely to find hands and manipulated objects~\cite{shan2020understanding}.
We only consider the object regions and select those with a confidence threshold greater than $0.01$ non-filtered by the typical non-maxima suppression operation of object detector pipelines~\cite{gkioxari2018detecting}.
Each region of interest is cropped from the original frame size, and resized such that the resolution of each object region is $112 
\times 112$.
During training, we enable the center and size jittering as means of data augmentation. The jittering is proportional to the size of the region of interest, and sampled from a uniform distribution $[1, 1.25)$.
For each frame, we consider three regions at most. At test time, we retained the most confident detected regions. Yet, we sampled object regions irrespective of the confidence score during training for data augmentation purposes.

Our \netname{} network uses a 2D-CNN backbone, ResNet-50~\cite{resnet2016}, to encode each object region.
Its Pool block corresponds to an integrated classifier followed by a aggregation module, the video prediction is done by classifying each object region independently, and then aggregating all the object prediction with a parameter-less Mean pooling operation.
We use the standard frame sampling~\cite{wang2018temporal}, and modestly consider 8 frames per video.
In contrast to prior art \cite{bulat2021space}, we did \underline{not} resort on additional test-time data augmentations (\eg\ multi-crop/views) per video. Thus, the performance of our model could improve further.

During the on-domain SOS self-supervised pre-training (\stagetwotrainname{}), we apply the standard set of photometric and geometric data augmentation (\eg, color jittering and cropping). Refer to~\cite{caron2020unsupervised} for more  details.
We resort in single-machine training with a batch size of 256 over 400 epochs and used the default optimization hyperparameters~\cite{caron2020unsupervised}.
We initialize the CNN backbone of \netname{} from an off-the-shelf self-supervised pre-trained network on ImageNet~\cite{caron2020unsupervised} (\ie, \stageonetrainname{} pre-training).
During the target task stage, we tune the whole \netname{} network end-to-end for the purpose of video classification. We train with SGD using an initial learning rate of $0.02$ and momentum $0.9$ with a batch size of 64 videos. We decay the learning rate by a factor of 0.1 after 20 epochs.
We implemented our approach using Pytorch~\cite{pytorch}, PytorchLightning~\cite{pytorchLightning} and the NUMFocus and image processing stack~\cite{numpy,pil}.

\subsection{Evaluation on EPIC-KITCHENS-100}
\label{sec:epic100_standard_eval}




\begin{figure}[t]
    \centering
    \includegraphics[width=\linewidth]{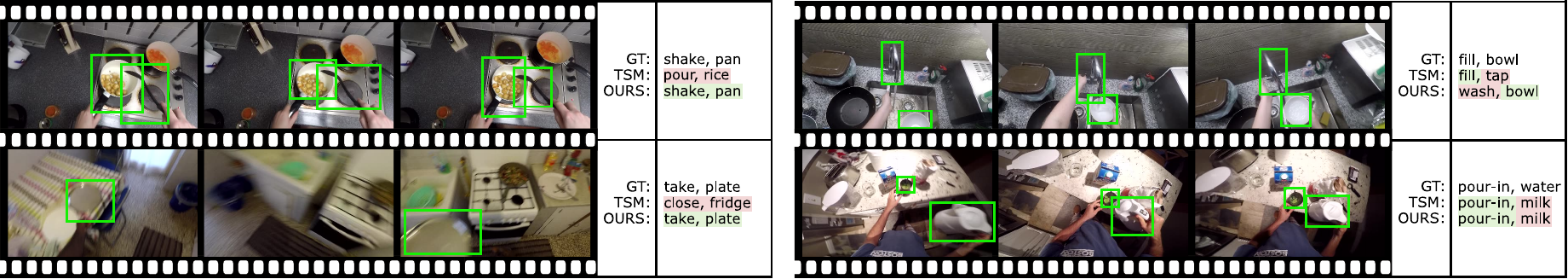}
    \vskip -0.3cm
    \caption{{\bf Qualitative results.} The green bounding boxes represent automatically detected object regions being manipulated. The left examples are corrected by our method, while the right two examples 
    either errors were maintained or flipped.}
    \label{fig:qualitative}
    \vspace{-0.5cm}
\end{figure}

\noindent {\bf Qualitative analysis }
Fig. \ref{fig:qualitative} depicts qualitative results, each quadrant depicts three frames from a particular video with its associated object regions, ground truth labels, and the predictions made by TSM as well as OURS (\ie, TSM+\netname{}). Correctly predicted verbs and nouns are highlighted as green, while incorrectly predicted as red. We observe that \netname{} network provides additional context to guide predictions correctly.
Fig. \ref{fig:qualitative} (bottom left) depicts an example where the noun of interest is mainly out of view during the duration of clip (only visible at the very beginning and at the end), hence the verb is also implicitly also not visible. Nevertheless, including the \netname{} module allows this information to be correctly captured and preserved.
Similarly, Fig. \ref{fig:qualitative} (top right) shows that TSM alone makes somewhat nonsensical prediction of ``fill, tap'', while ours predicts ``wash, bowl" which is arguably visually and semantically closer to the ground truth ``fill, bowl'', after all filling a bowl looks much like washing a bowl. However, both TSM and ours still struggle with visually confusing noun classes, as shown in Fig. \ref{fig:qualitative} (bottom right). Although both models correctly predict the verb ``pour-in", they both confuse the white water kettle with a jug of milk. Perhaps surprisingly, we appreciate that TSM often struggles with seemingly simple cases (noun and verb clearly visible and executed), as shown in Fig. \ref{fig:qualitative} (top left), while the inclusion of \netname{} greatly alleviates this problem.

\noindent {\bf Baselines }
For extensive comparative evaluation, we consider a variety of state-of-the-art action recognition models
including (1) top CNN action models:
TSM \cite{lin2019temporal}, and
SlowFast  \cite{feichtenhofer2019slowfast}
designed for generic action recognition tasks
with varying architectural design for spatio-temporal representation learning,
(2) a recent vision transformer: X-ViT \cite{bulat2021space} with superior cost-effective formulation for spatio-temporal attention learning,
and 
(3) the latest egocentric action model: IPL
\cite{wang2021interactive}
designed to learn superior active object representation
subject to human motion cues.
We test the effective of our \netname{} in improving several above methods with the proposed fusion method.

\begin{table*}[!t]
  \centering
  \resizebox{\columnwidth}{!}{%
   \setlength{\tabcolsep}{3pt}
  \begin{tabular}{l|lc|ccc|ccc|ccc|ccc}
    \toprule
    \multirow{3}{*}{Modality} & \multirow{3}{*}{Method} & \multirow{3}{*}{TTDA} & \multicolumn{6}{c|}{Overall} & \multicolumn{3}{c|}{Unseen participants} & \multicolumn{3}{c}{Tail classes} \\
    & & & \multicolumn{3}{c|}{Top-1} & \multicolumn{3}{c|}{Top-5} & \multicolumn{3}{c|}{Top-1} & \multicolumn{3}{c}{Top-1} \\
    & & & Verb & Noun & Action & Verb & Noun & Action & Verb & Noun & Action & Verb & Noun & Action \\
    \midrule
    V & TSM$^\dagger$ \cite{lin2019temporal} & \xmark
    & 63.0 & 47.3 & 35.6
    & 88.9 & 74.2 & 57.2 
    & \bf 54.6 & 37.1 & 26.3
    & 35.9 & 26.7 & 18.1
    \\
    V & + {\bf \netname} & \xmark
    & \bf 64.0 & \bf 52.3 & \bf 39.0
    & \bf 90.2 & \bf 78.4 & \bf 61.6
    &     53.1 & \bf 40.3 & \bf 27.2
    & \bf 36.2 & \bf 29.9 & \bf 20.5
    \\
    \cmidrule(lr{1em}){2-15}
    V & SlowFast$^\dagger$ \cite{feichtenhofer2019slowfast} & \xmark
    & 65.6     & 50.0 & 38.5 
    & \bf 90.0 & 75.6 & 58.7
    & \bf 56.7 & 41.6 & 30.0
    & 36.3     & 23.3 & 18.7
    \\
    V & +{\bf \netname} & \xmark
    & \bf 66.5 & \bf 53.5 & \bf 40.9
    & 89.6     & \bf 77.8 & \bf 61.8
    & 56.6     & \bf 43.6 & \bf 31.1
    & \bf 37.6 & \bf 27.3 & \bf 20.3
    \\
    \cmidrule(lr{1em}){2-15}
    V & X-ViT$^\dagger$ \cite{bulat2021space} & \xmark
    & 62.9 & 49.1 & 37.7
    & 87.5 & 73.8 & 57.1
    & 54.4 & 41.7 & 30.5
    & 35.7 & 28.0 & 19.42
    \\
    V & +{\bf \netname} & \xmark
    & \bf 64.7 & \bf 54.6 & \bf 40.9
    & \bf 91.0 & \bf 79.3 & \bf 62.8
    & \bf 54.6 & \bf 43.8 & \bf 31.2
    & \bf 37.0 & \bf 31.1 & \bf 21.2
    \\
    \cmidrule(lr{1em}){2-15}
    V & X-ViT \cite{bulat2021space}$^\dagger$ & \cmark 
    & 68.7 & 56.4 & 44.3
    & 90.4 & 79.4 & 64.5
    & 57.8 & 47.4 & 34.9
    & 38.2 & 31.8 & 22.4 
    \\
    V & +{\bf \netname} & \cmark
    & \underline{\bf 69.3} & \underline{\bf 57.9} & \underline{\bf 45.7}
    & \bf 91.2             & \underline{\bf 81.1} & \underline{\bf 66.2}
    & \bf 57.9             & \underline{\bf 49.3} & \underline{\bf 36.3}
    & \underline{\bf 38.6} & \underline{\bf 33.1} & \underline{\bf 23.7}
    \\
    \midrule
    V+F & TSM$^\dagger$ \cite{lin2019temporal} & \xmark
    & 67.9 & 49.1 & 38.4
    & 91.0 & 74.9 & 60.7
    & 58.5 & 39.2 & 29.3
    & 36.8 & 22.5 & 17.7
    \\
    V+F & +{\bf \netname} & \xmark
    & \bf 68.4             & \bf 53.7 & \bf 41.8
    & \underline{\bf 91.4} & \bf 78.7 & \bf 64.0
    & \underline{\bf 58.7} & \bf 42.7 & \bf 30.8
    & \bf 37.1             & \bf 26.7 & \bf 20.1
    \\
    \cmidrule(lr{1em}){2-15}
    V+F & IPL(I3D) \cite{wang2021interactive} & \xmark
    &  67.82 & 50.87 & 39.87 & - & - & - & - & - & -  & - & - & -
    \\
    V+F & IPL(R2+1D) \cite{wang2021interactive} & \xmark
    & 68.61 & 51.24 & 40.98 & - & - & - & - & - & -  & - & - & - \\
    \bottomrule
  \end{tabular}
  } 
  \caption{{\bf Results on the validation set of EPIC-KITCHENS-100} \cite{damen2020rescaling}. 
  Modality: V=Visual; F=Optical flow. 
  $^\dagger$: Results computed with the model weights released by the authors of \cite{damen2020rescaling,bulat2021space}.
  TTDA: Test-time data-augmentation (\eg, multi-crop).
  Underlined numbers correspond to best results across the board. Bold numbers highlight the best between a proxy model and proxy + ours.
  All in all, our model yields state-of-the-art results and it is versatile as improved four action classification models.
  } 
  \label{tab:sota_epic}
  \vspace{-25pt}
\end{table*}

\noindent {\bf Results analysis }
We report the action recognition results in Table \ref{tab:sota_epic}.
We have the following observations:
{\bf (1)} The performance of CNN action models,
SlowFast and TSM, is similar to the recently introduced X-ViT without heavy test-time data augmentation.
{\bf (2)} Importantly, our \netname{} further improves SlowFast(V), TSM(V+F) and X-ViT
by 2.4\%, 3.4\%, and 3.2\% on overall action accuracy, respectively.
Without using computationally expensive optical flow, TSM still benefits from \netname{} at similar scale.
These results verify the versatility and consistent usefulness of our \netname{} in enhancing prior art models.
It is also seen that the improvement is mainly achieved on noun recognition as expected. 
This indicates that these methods are limited in extracting active objects
in manipulation and interaction with human hands from cluttered background and scene.
This is exactly the motivation of learning our \netname{} model. 
{\bf (3)}
We also observes that our \netname{} clearly improves the accuracy scores on unseen participants and tail classes. 
This implies that exploiting our \netname{} could help reduce the negative impact of domain shift (seen vs. unseen participants in this case) and mitigate the overwhelming effect from head classes to tail classes, concurrently.
{\bf (4)}
Along with TSM, our method also clearly outperforms the latest
egocentric action model IPL which is designed to
learn better active object representations.
This indicates that without explicit region detection,
the CNN model itself is less effective to isolate the active objects
from the scenes.
{\bf (5)}
The recent X-ViT model  significantly lifts the performance of CNN by means of additional test-time data augmentation (\ie, averaging the results from 3 crops per video).
It is worth noting that even after triplicating the computational budget, our computationally modest \netname{} representation stills provides a further gain of 1.4\% on overall actions to this model.
It is possible that our \netname{} and previous CNNs model could benefit from the test-time data augmentation techniques. If so, it is worth balancing the euphoria surrounding Transformer-based models with fair and rigorous experimental evaluation.

\vspace{-10pt}
\subsection{Evaluation on EGTEA}

\begin{table}[!t]
  \centering
  \setlength{\tabcolsep}{5pt}
  \resizebox{0.6\columnwidth}{!}{%
  \begin{tabular}{l|l|ccc|c}
    \toprule
   {Modality} & {Method} & Split 1 & Split 2 &Split 3 & Avg \\
    \hline
    V & TSM \cite{lin2019temporal} & 61.2 & 61.2 & 59.6 & 60.7
    \\
    V & TSM+{\bf \netname} & \bf 62.0 &\bf  61.9 &\bf 60.0 &\bf 61.3
    \\
    \midrule
    V+F & 2SCNN* \cite{simonyan2014very} &
    43.8 & 41.5 & 40.3 & 41.8
    \\
   V+F & I3D*  \cite{carreira2017quo} &
   55.8 & 53.1 & 53.6 & 54.2
   \\
   V+F & I3D+Center Prior* &
   52.5 & 49.4 & 49.3 & 50.4
   \\
   V+F & I3D+EgoConv* \cite{singh2016first} &
   54.2 & 51.5 & 49.4 & 51.7
   \\ 
   V+F & Ego-RNN-2S* \cite{sudhakaran2018attention} &
   52.4 & 50.1 & 49.1 & 50.5
   \\
   V+F & LSTA-2S* \cite{sudhakaran2019lsta} &
   53.0 & - & - & -
   \\
   V+F & MutualCtx-2S* \cite{huang2020mutual} &
   55.7 & - & - & -
   \\
   V+F & Prob-ATT \cite{li2021eye} &
   56.5 & 53.5 & 53.6 & 54.5
   \\
   V+F & I3D+IPL \cite{wang2021interactive} &
   60.2 & 59.0 & 57.9 & 59.1
   \\
    \midrule
    V+F+G & I3D*  \cite{carreira2017quo} &
    53.7 & 50.3 & 49.6 & 51.2
    \\ 
    V+F+G &  Prob-ATT \cite{li2021eye} &
    57.2 & 53.8 & 54.1 & 55.0
    \\
    \bottomrule
  \end{tabular}
  } 
  \caption{
  {\bf Comparison against state of the art in EGTEA \cite{li2018eye}}. 
  Modality: V=Visual; F=Optical flow; G=Gaze. 
  *: Results taken from \cite{li2021eye}.}
  \label{tab:sota_egtea}
  \vspace{-15pt}
\end{table}

\noindent {\bf Baselines }
Compared to EPIC-KITCHENS-100, this dataset is uniquely featured with 
eye gaze tracking information. We compare our method with the following alternatives: 
(1) 2SCNN \cite{simonyan2014very}: A two stream networks with VGG16
backbone, 
(2) I3D \cite{carreira2017quo}:  A two stream I3D with joint
training of RGB and optical flow, 
(3) I3D+Gaze: Instead of average pooling, the ground
truth human gaze is leveraged to pool the features from the
last conv layer,
(4) I3D+Center Prior: A fixed 2D Gaussian distribution at the center is applied
as the attention map for pooling the last conv features,
(5) I3D+EgoConv \cite{singh2016first}: Adding a stream of egocentric cues
for encoding head motion and hand mask, and the output scores are late fused with RGB and optical flow streams, 
(6) Ego-RNN-2S \cite{sudhakaran2018attention} and LSTA-2S \cite{sudhakaran2019lsta}:
two recurrent networks with soft-attention,
(7) MutualCtx-2S \cite{huang2020mutual}: A gaze enhanced action model 
trained by alternating between gaze estimation
and action recognition,
(8) Prob-ATT \cite{li2021eye}: A state-of-the-art
joint gaze estimation and action recognition model featured with a stochastic gaze distribution formulation,
(9) TSM \cite{lin2019temporal}:
A recent strong action recognition model with efficient temporal shift operation between nearby frames for motion modeling,
(10) IPL \cite{wang2021interactive}:
A recent state-of-the-art egocentric action model
as discussed earlier.
We combine our \netname{} with TSM in evaluating this dataset.

\paragraph{Results analysis.}
Table \ref{tab:sota_egtea} reports the results.
We observed that:
{\bf (1)} using gaze modality does not guarantee 
superior results; For example, without gaze IPL still achieves better results than Prob-ATT using gaze. This suggests that video data alone already provide rich information and the key is how to learn and extract discriminative action information with proper model design. %
The way to leverage the gaze data is also equally critical.
{\bf (2)} Similarly, the computationally expensive optical flow is not the highest performance promise, as indicated by the excellent performance with TSM using only 2D video frames as input.
{\bf (3)} Importantly, our \netname{} again further improves the performance of TSM consistently over all the splits, suggesting the generic efficacy of our method on
a second test scenario.

\vspace{-5pt}
\subsection{SOS and Long-tail Learning}
\label{exp:sos_lt}

Here we present a rigorous assessment on the role of our SOS pre-training for dealing with class-imbalance distributions in videos.
\textbf{\textit{Dataset and metrics}}. We perform the study in the validation set of EPIC-KITCHENS-100 using the standard metrics for evaluating long-tail learning algorithms \cite{dong2018imbalanced}, class-balanced average accuracy for verb, noun and action. Concretely, each video is weighted by the inverse of the number of instances of its corresponding label. Note that we retain the tail classes definition from \cite{damen2020rescaling}, but the metrics reported in this section differ from those in Table \ref{tab:sota_epic}.
\textbf{\textit{Baselines}}. We consider our \netname{} network aided (or not) with our \sslshortname{} pre\=/training and using (or not) the long-tail (LT) loss of \cite{menon2020long} during the target task fine-tuning. We also report of single-crop X-ViT \cite{bulat2021space}, using the pre-trained weights of the authors, and our fused model.

\begin{table*}[!t]
  \centering
  \setlength{\tabcolsep}{4pt}
  \begin{tabular}{llcc|ccc|ccc}
    \toprule
    & \multirow{2}{*}{Method} & \multirow{2}{*}{SOS} & \multirow{2}{*}{\shortstack{LT loss\\ \cite{menon2020long}}}
    & \multicolumn{3}{c|}{Overall} & \multicolumn{3}{c}{Tail classes} \\
    & & & & Verb & Noun & Action & Verb & Noun & Action \\
    \midrule
    \bf A & {\bf \netname} &\xmark &\xmark
    & 18.2 & 21.1 & 8.7
    & 13.2 & 11.9 & 6.1
    \\
    \bf B & {\bf \netname} & \xmark & \cmark
    & 27.5 & 25.0 & 8.1 
    & 25.4 & 19.9 &	7.1
    \\
    \bf C & {\bf \netname} & \cmark & \xmark
    & 20.8 & 25.0 & \bf 10.8 
    & 15.9 & 16.7 & 8.6
    \\
    \bf D & {\bf \netname} & \cmark & \cmark
    & \underline{\bf 30.3} & \bf 28.0 & 9.4
    & \underline{\bf 28.6} & \bf 23.4 & \bf 8.7
    \\
    \midrule
    \bf E & X-ViT & \xmark & \xmark
    & 22.1 &	25.9 & 12.8
    & 15.9 &	17.8 &	9.5
    \\
    \bf F & X-ViT + {\bf \netname} & \cmark & \cmark
    & \bf 30.2 & \underline{\bf 31.8} & \underline{\bf 15.2}
    & \bf 25.2 & \underline{\bf 24.9} & \underline{\bf 12.8}
    \\
    \bottomrule
  \end{tabular}
  \caption{{\bf Long-tail results (accuracy) on the validation set of EPIC-KITCHENS-100}, showcasing the impact of SOS for dealing with class-imbalance distributions. Refer to text for details about the metric.
  } 
  \label{tab:lt_epic}
  \vspace{-25pt}
\end{table*}

\paragraph{Results analysis.} Table \ref{tab:lt_epic} reports the results.
We observed that:
{\bf (1)} Our \sslshortname{} pre-training indeed helps for dealing with class-imbalanced scenarios. It naturally improves noun and action classes the most, by +3.9\% and +2.1\% overall classes respectively (row \textbf{C} v.s. \textbf{A}). Note that  aiding \netname{} with \sslshortname{} (\textbf{C}) significantly reduced the gap between the X-ViT (\textbf{E}) and our plain \netname{} (\textbf{A}).
{\bf (2)} \sslshortname{} is complementary to the state-of-the-art LT learning approach. \sslshortname{} and the LT loss \cite{menon2020long} yields the best results for the \netname{} model, except for overall action classes where \sslshortname{} pre-training achieves the best result. The relatively minor setback evidences the relevance of studying LT and SSL pre-training within the context of structured labels such as actions in videos.
{\bf (3)} Our fused model (\textbf{F}) , X-ViT with \netname{} aided with \sslshortname{} and the LT loss \cite{menon2020long}, achieves the state-of-the-art on LT accuracy for noun and action by a large margin.

Overall, we have validated the relevance and complementary of \sslshortname{} for dealing with long-tail learning scenarios in egocentric action recognition.





\vspace{-5pt}
\subsection{Ablation Study}
\label{sec:exp_ablation}
As prior art~\cite{wang2021interactive}, we validate the contribution of the major components of our approach on EPIC-KITCHENS-100. We report the top-1 accuracy for Verb, Noun and Action for overall and tail classes following~\cite{damen2020rescaling}.

\paragraph{Impact of an off-the-shelf self-supervised encoder.}
Encoders learned via supervised learning might overfit the label space of the pre-training source. In contrast, state-of-the-art self-supervised learning relies on a set of data augmentations and measure of instance similarity.
Thus, it is worth studying whether an off-the-shelf self-supervised encoder offers any advantage for representing the visual regions of handled objects in a target dataset.
We use two different off-the-shelf encoders learned via: (A) supervision, and (B) self-supervision, both using ImageNet~\cite{russakovsky2015imagenet} and acting as initialization for our \netname{} backbone.
Table~\ref{tab:single_table_ablation} row {\bf A} reports the results of using a supervised pre-trained encoder as initialization, while row {\bf B} reports when initialization is done via self-supervision.
We observe that self-supervised initialization improves the performance \wrt\ supervised initialization in noun by 1.3\% and action by 0.5\% without significantly degrading verb performance.
These results echo the relevance and popularity of self-supervised pre-training~\cite{goyalCLXWPSLMJB2021} in the domain of egocentric visual perception, which has been relatively under-explored.

\begin{table}[!t]
\centering
\setlength{\tabcolsep}{3pt}
\begin{tabular}{llc|ccl|ccl}
\toprule
\multirow{2}{*}{} & \multirow{2}{*}{Method} & \multirow{2}{*}{OD} & \multicolumn{3}{c|}{Overall} & \multicolumn{3}{c}{Tail classes} \\
& & & Verb & Noun & Action
& Verb & Noun & Action \\
\midrule
{\bf A} & Supervised pre-training & \xmark
& 49.3 & 44.5 & 27.4 & 32.0 & 23.6 & 14.8 \\
{\bf B} & SwAV & \xmark
& 49.2 & 45.8 & 27.9 & 31.2 & 21.2 & 14.0 \\
{\bf C} & SwAV & \cmark
& 49.9 & 47.0 & 28.7 & 30.6 & 23.1 & 14.6 \\
{\bf D} & SwAV-S & \cmark
& \bf 51.5 & \bf 48.5 & {\bf 30.2} & \bf 32.4 & \bf 25.6 & {\bf 16.4} \\
\bottomrule
\end{tabular}
\caption{
\label{tab:single_table_ablation}
{\bf Ablation of different components of our approach on EPIC-KITCHENS-100.}
Rows {\bf A}-{\bf D} report the top-1 accuracy of the \netname{} model by itself (\ie, without video network fusion) for \textit{overall} and \textit{tail} Verb, Noun and Action instances~\cite{damen2020rescaling}.
OD: On-Domain pre-training in target dataset, EPIC-KITCHENS-100.
}
\vspace{-25pt}
\end{table}

\paragraph{Impact of self-supervised adaptation.}
We gauge the impact of adapting the off-the-shelf self-supervised representation to the domain of interest, EPIC-KITCHENS-100 manipulated object regions.
For this purpose, we employ the standard SwAV loss (Eq.~\eqref{eq:swav_loss}) during the \stagetwotrainname{} pre-training, (\cf\ Fig. \ref{fig:pipeline}). We use the original set of data augmentations and optimization hyper-parameters as in~\cite{caron2020unsupervised}.
Table~\ref{tab:single_table_ablation} row {\bf C} reports the results of the self-supervised domain adaptation with SwAV loss over individual object regions (Fig. \ref{fig:swavs} - top).
Adapting the representation helps to boost the predictive power for verb, noun and action by 0.7\%, 1.2\% and 0.8\% \wrt no domain adaptation Table~\ref{tab:single_table_ablation} row {\bf B}.

\paragraph{Relevance of SwAV-S.}
We validate the impact of SwAV-S, which treats actions as a natural self-supervised transformation.
For this purpose, we employ the SwAV-S loss (Eq.~\eqref{eq:swavs_loss}) during the \stagetwotrainname{} pre-training (see Fig. \ref{fig:pipeline}). We kept the same set of hyper-parameters as SwAV.
Table~\ref{tab:single_table_ablation} row {\bf D} reports the results of SwAV-S.
We observe a more significant boost in performance across the board
and surpasses SwAV (Table~\ref{tab:single_table_ablation} 
row {\bf C}) 
by 1.6\% for verb, 1.5\% for noun and 1.5\% for action. 
This result validates the stronger benefit of our novel approach for self-supervised domain adaptation.

\section{Conclusions}

We have presented a novel method for incorporating object information into egocentric action recognition models. Resorting to self-supervised learning (SSL), our method does not require object-level annotations, 
and exploits hand-object contact detection to directly select foreground objects and sieve out background distractors. Through a careful SSL design, it obtains a strong representation for objects being manipulated.
Experiments show that prior state-of-the-art video models can consistently benefit from our object model on two egocentric datasets, with improved ability to tackle the challenging class imbalance.


\clearpage

\section{Appendix}
We complement the manuscript with the following items:

\begin{itemize}
    \item A video, {\bf with voice-over of one author}, summarizing our work. Feel free to refer to it for a brief summary of our work and appreciate qualitative results over videos.
    The codec of the video is H264 - MPEG-4 AVC with resolution $1280 \times 720$ and frame rate of 30 fps. The following media players work well for us: VLC media player \footnote{\url{https://www.videolan.org/vlc/index.html}}; Google Chrome, simply drag and drop it; Microsoft Movies \& TV; and Windows Media Player.
    
    
    \item Results in the test set of EPIC-KITCHENS-100 without public labels,
    refer to Section \ref{sec_a:test_epic100}.
    
    \item Additional results of two model ensemble ablations,
    refer to Section \ref{sec_a:ensemble_ablation}.
    
    \item Additional qualitative results,
    refer to Section \ref{sec_a:qualitative_results}.
    
    \item A pragmatic take to understand quantitative differences in Epic-Kitchens 100~\cite{damen2020rescaling},
    refer to Section \ref{sec_a:tolerance}.
    
    
\end{itemize}

\subsection{Results in EPIC-KITCHENS-100 test set}
\label{sec_a:test_epic100}

To validate the generalization of the insights presented on the main paper, we resort in the test set of EPIC-KITCHENS-100 with sequestered labels.
Concretely, we submitted the predictions of a given model onto the evaluation server.
Table \ref{tab:epic_test} reports the results from the evaluation server, with the metrics established by the dataset creators \cite{damen2020rescaling}.

Overall the observations drawn from the validation can be seen in the test set with sequestered data. We repeat some of the main observations here:
{\bf (1)} The performance of CNN action models, TSM, is similar to the recently introduced X-ViT without heavy test-time data augmentation.
{\bf (2)} Importantly, our \netname{} further improves X-ViT
by 1.6\%, 5.4\%, and 2.6\% on overall verb, noun and action accuracy, respectively.
It is also seen that the improvement is mainly achieved on noun recognition as expected. 
Thus, it indicates that the attention mechanism of transformer is limited in extracting active objects
in manipulation and interaction with human hands from cluttered background and scene.
This is exactly the motivation of learning our \netname{} model. 
{\bf (3)}
We also observes that our \netname{} clearly improves the accuracy scores on unseen participants and tail classes. 
This implies that exploiting our \netname{} could help reduce the negative impact of domain shift (seen vs. unseen participants in this case) and mitigate the overwhelming effect from head classes to tail classes, concurrently.
{\bf (4)}
The recent X-ViT model  significantly lifts the performance of CNN by means of additional test-time data augmentation (\ie, averaging the results from 3 crops per video).
It is worth noting that even after triplicating the computational budget, our computationally modest \netname{} representation stills provides a further gain of 1.2\% on overall actions to this model.

\begin{table*}[!t]
  \centering
  \resizebox{\columnwidth}{!}{%
   \setlength{\tabcolsep}{3pt}
  \begin{tabular}{l|lc|ccc|ccc|ccc|ccc}
    \toprule
    \multirow{3}{*}{Modality} & \multirow{3}{*}{Method} & \multirow{3}{*}{TTDA} & \multicolumn{6}{c|}{Overall} & \multicolumn{3}{c|}{Unseen participants} & \multicolumn{3}{c}{Tail classes} \\
    & & & \multicolumn{3}{c|}{Top-1} & \multicolumn{3}{c|}{Top-5} & \multicolumn{3}{c|}{Top-1} & \multicolumn{3}{c}{Top-1} \\
    & & & Verb & Noun & Action & Verb & Noun & Action & Verb & Noun & Action & Verb & Noun & Action \\
    \midrule
    V & X-ViT$^\dagger$ \cite{bulat2021space} & \xmark
    & 58.5 & 46.9 & 34.5
    & 85.2 & 71.4 & 53.0
    & 50.4 & 37.7 & 25.7
    & 28.2 & 21.9 & 14.0
    \\
    V & +{\bf \netname} & \xmark
    & \bf 60.1 & \bf 52.3 & \bf 37.1
    & \bf 87.7 & \bf 77.7 & \bf 59.0
    & \bf 53.0 & \bf 44.9 & \bf 28.8
    & \bf 30.4 & \bf 25.2 & \bf 16.1
    \\
    \cmidrule(lr{1em}){2-15}
    V & X-ViT \cite{bulat2021space}$^\dagger$ & \cmark
    & 64.3 & 52.9 & 40.3
    & 88.1 & 77.4 & 60.6
    & 58.2 & 45.7 & 32.0
    & 31.2 & 26.3 & 17.7
    \\
    V & +{\bf \netname} & \cmark
    & \underline{\bf 64.7} & \underline{\bf 55.0} & \underline{\bf 41.5}
    & \underline{\bf 89.2} & \underline{\bf 79.5} & \underline{\bf 62.7}
    & \underline{\bf 58.6} & \underline{\bf 48.0} & \underline{\bf 33.0}
    & \underline{\bf 31.4} & \underline{\bf 27.6} & \underline{\bf 18.1}
    \\
    \midrule
    V & TSM$^\dagger$ \cite{lin2019temporal} & \xmark
    & 58.5 & 45.9 & 32.8
    & 86.4 & 72.3 & 54.0
    & 53.1 & 41.1 & 27.1
    & 27.9 & 21.5 & 13.8
    \\
    F & TSM$^\dagger$ \cite{lin2019temporal} & \xmark
    & 60.9 & 33.8 & 26.2
    & 86.1 & 57.6 & 43.5
    & 55.5 & 27.5 & 20.1
    & 26.0 & 6.4  & 7.5
    \\
    \bottomrule
  \end{tabular}
  } 
  \caption{{\bf Results on the test set of EPIC-KITCHENS-100} \cite{damen2020rescaling} with models \underline{only trained in the training set}. 
  Modality: V=Visual; F=Optical flow. 
  $^\dagger$: Results computed with the model weights released by the authors of \cite{damen2020rescaling,bulat2021space}.
  TTDA: Test-time data-augmentation (\eg, multi-crop).
  Underlined numbers correspond to best results across the board. Bold numbers highlight the best between a proxy model and proxy + ours.
  All in all, our model complements X-ViT and yields state-of-the-art results.
  } 
  \label{tab:epic_test}
\end{table*}

\subsection{Additional ablation of two models ensemble}
\label{sec_a:ensemble_ablation}
Our full video prediction network corresponds to the fusion of our \netname{} Net and one standard video classification network, \cf\ main paper Sec. \textcolor{red}{3.2} and Fig \textcolor{red}{4}.
The intuition behind the proposed fusion are: (1) 
it allows to 
validate the relevance of handled objects and learning a specialized neural network representation for them. (2) It is a simple way to integrate the information capture by our \netname{} Net without the need of model re-training. 
(3) It combines the complementary information discovered by our \netname{} Net relating to objects while the video net provides the scene context.
For fair comparison, we compare the results of our fused model \wrt\ the proxy video network used in our fusion, and the ensemble of two networks of a given proxy model. 

\paragraph{\underline{\normalfont{Implementation details.}}} We used the standard TSM video architecture \cite{lin2019temporal} with the pre-trained networks given by \cite{damen2020rescaling} as the proxy video network. For the two model ensemble baseline, we train an additional TSM model using the code and training recipe given by \cite{damen2020rescaling}. For our fused model, we followed the training pipeline presented in the main paper.
Table \ref{tab:epic_ensemble} reports the experimental results in the validation set of EPIC-KITCHENS-100 with the metrics established by the dataset creators \cite{damen2020rescaling}.
We have the following observations:
(1) TSM ensemble yield an absolute performance improvement of +0.7\%, +1.0\% and +0.6\% on overall verb, noun and action respectively. Meanwhile, our model fusion with \netname{} yields significantly better absolute improvement for noun and action (+5.0\% and +3.4\%), and slightly better absolute improvement for verb (+1.0\%).
(2) We also observes that our \netname{} clearly improves the accuracy scores on unseen participants (for noun and action) and tail classes (for verb, noun and action). As expected, the most significant performance improvement are from noun accuracy.

In summary, we have validated the significance of the results from our fused model, with \netname{}, in comparison to a stronger baseline.

\begin{table*}[!t]
  \centering
  \resizebox{\columnwidth}{!}{%
   \setlength{\tabcolsep}{3pt}
  \begin{tabular}{l|ccc|ccc|ccc|ccc}
    \toprule
     \multirow{3}{*}{Method} & \multicolumn{6}{c|}{Overall} & \multicolumn{3}{c|}{Unseen participants} & \multicolumn{3}{c}{Tail classes} \\
    & \multicolumn{3}{c|}{Top-1} & \multicolumn{3}{c|}{Top-5} & \multicolumn{3}{c|}{Top-1} & \multicolumn{3}{c}{Top-1} \\
    & Verb & Noun & Action & Verb & Noun & Action & Verb & Noun & Action & Verb & Noun & Action \\
    \midrule
    TSM$^\dagger$ \cite{lin2019temporal}
    & 63.0 & 47.3 & 35.6
    & 88.9 & 74.2 & 57.2 
    & 54.6 & 37.1 & 26.3
    & 35.9 & 26.7 & 18.1
    \\
    TSM ensemble
    & 63.7 & 48.3 & 36.2
    & 89.7 & 75.3 & 58.6
    & \bf 55.3 & 38.5 & 26.8
    & 34.7 & 26.3 & 18.0
    \\
    TSM + {\bf \netname}
    & \bf 64.0 & \bf 52.3 & \bf 39.0
    & \bf 90.2 & \bf 78.4 & \bf 61.6
    &     53.1 & \bf 40.3 & \bf 27.2
    & \bf 36.2 & \bf 29.9 & \bf 20.5
    \\
   \bottomrule
  \end{tabular}
  } 
  \caption{{\bf Results on the validation set of EPIC-KITCHENS-100} \cite{damen2020rescaling}. 
  $^\dagger$: Results computed with the model weights released by the authors of \cite{damen2020rescaling}.
  Bold numbers highlight the best between a proxy model (TSM), two model ensemble of the proxy model (TSM ensemble), and proxy + ours (TSM + \netname{}).
  All in all, our \netname{} improves further the proxy model validating the relevance of our handled objects representation.
  } 
  \label{tab:epic_ensemble}
\end{table*}

\begin{figure*}[!t]
    \centering
    \includegraphics[width=\linewidth]{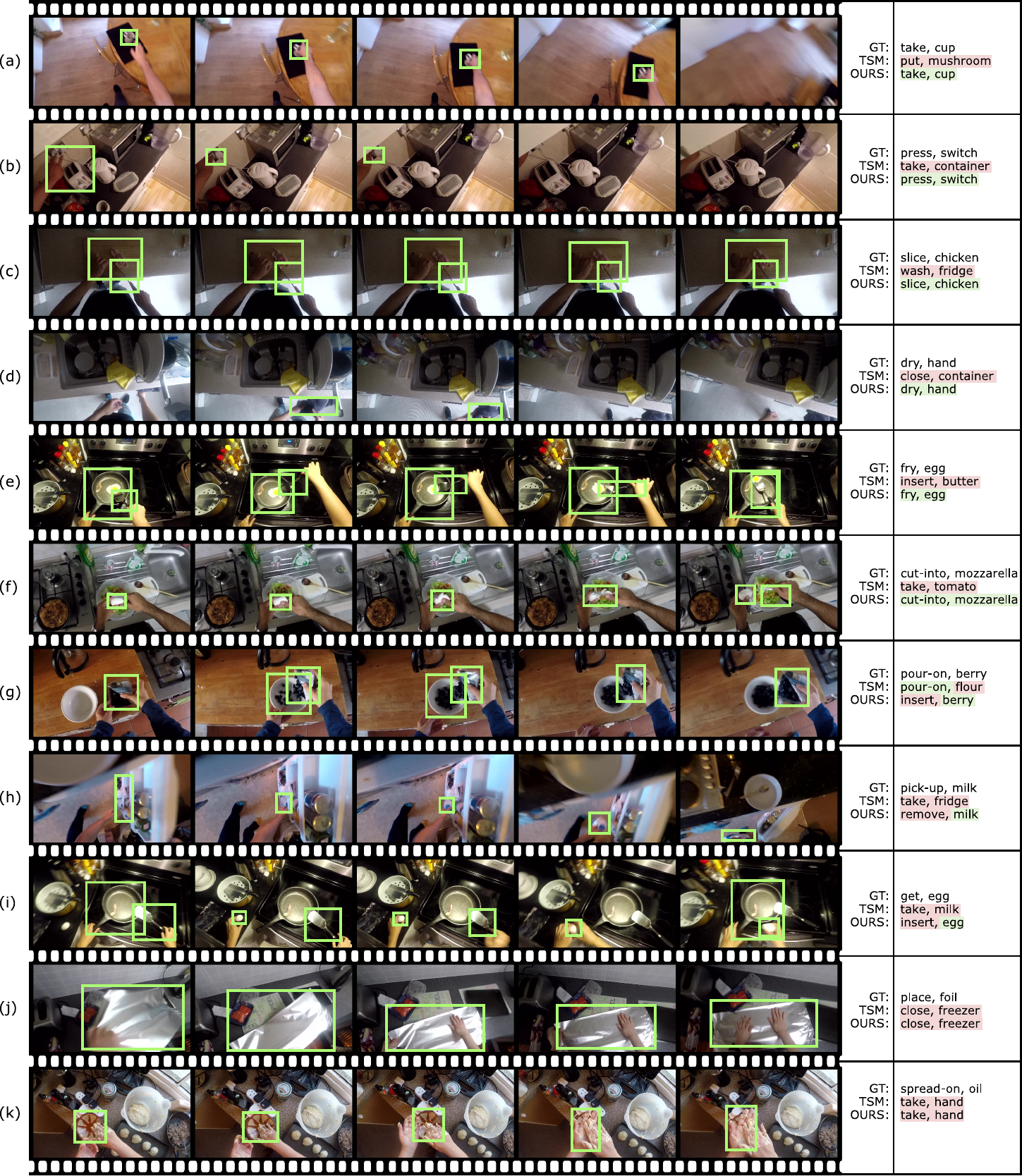}
    \caption{
    {\bf Qualitative results.} The green bounding boxes represent automatically detected object regions being manipulated. Rows ({\it a - f}) represent examples that are corrected by our method, rows ({\it g - i}) examples that are partially correct, while rows ({\it j - k}) depicts failure cases for both methods.
    }
    \label{fig:qualitative_sup}
\end{figure*}

\subsection{Qualitative results}
\label{sec_a:qualitative_results}

Fig.~\ref{fig:qualitative_sup} depicts additional qualitative results to those shown in Fig.~\textcolor{red}{5} in the \cf\ main manuscript. Here, each row depicts five evenly spaced frames from a particular example with its associated object regions, the ground truth labels, predictions made by TSM and predictions made by TSM+\netname{} (ours). Same as before, correctly predicted verbs and nouns are highlighted as green, while incorrectly predicted as red. These additional results further reinforce the notion that the additional context provided by the \netname{} module guides prediction correctly. Fig. \ref{fig:qualitative_sup} ({\it a - f}), depicts examples where the noun of interest is either very small ({\it a}, {\it b}, {\it e} and {\it f}), out of view during a large portion of the clip ({\it b} and {\it d}) or not well illuminated ({\it c}). Nevertheless, including the \netname{} module allows this information to be correctly captured and preserved, same which is lost or ignored with pure TSM. Fig. \ref{fig:qualitative_sup} , ({\it g}) depicts an example where both TSM and ours make a wrong prediction, in the case of TSM the verb is correctly identified, however, the noun predicted is very different from the ground truth. In this same example, our method correctly predicts the noun, while incorrectly predicting the verb; however, one could argue that the verb wrongly predicted by our method could apply to the example, while the same cannot be said by the wrongly predicted noun by TSM. Similar arguments as the ones just discussed for Fig.~\ref{fig:qualitative_sup} ({\it g}) would apply to {\it h}, where additionally TSM predictions of both verb and noud are incorrect.
Finally, Fig.~\ref{fig:qualitative_sup} ({\it j} - {\it k}), shows two fail examples for both methods, where despite the fact that our method makes use of object boxes largely associated with the target noun, the proposed method still fails.

\begin{figure*}[!t]
    \centering
    \includegraphics[width=\linewidth]{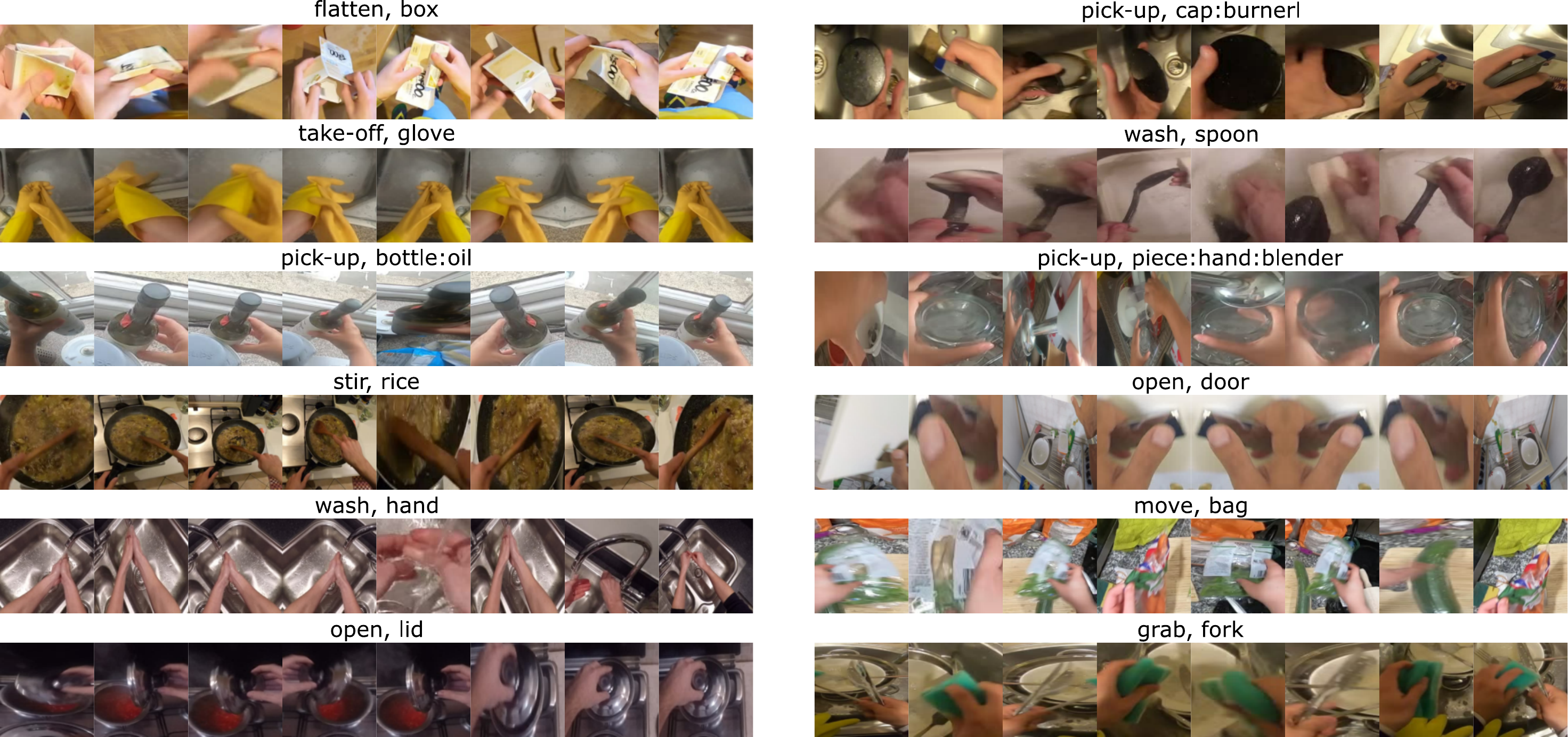}
    \caption{
    {\bf Qualitative results.} SwAV-S input objects and their related \textit{real} spatio-temporal transformations.
    }
    \label{fig:qualitative_swav_patches_sup}
\end{figure*}

Fig.~\ref{fig:qualitative_swav_patches_sup} shows examples of patches representing objects used during \sslshortname{} pre-training stage, \ie\  to train our proposed SwAV-S. The figure also displays the associated labels from the video for reference, but it is worth noting that our approach does not exploit the given labels during the \sslshortname{} pre-training. There, it can be appreciated the \textit{real} spatio-temporal transformation that objects experience. These, represent the sets of objects presented to SwAV-S for cluster assignment and prediction. Fig.\ref{fig:qualitative_swav_patches_sup} left column  illustrates cases where the action-process can easily be seen as the primary source of variation (\eg, the first row exhibits large deformations of a box being flattened), while the right column illustrates more challenging examples, that nonetheless, are able to capture the semantics of the underlying action. The examples presented in Fig.\ref{fig:qualitative_swav_patches_sup} left, also elucidates why our approach still works without the need of tracking or strictly enforcing object correspondences across frames.

\subsection{Performance improvements perspective}
\label{sec_a:tolerance}

Recently, a study on the influence of random seeds in modern deep learning architectures in computer vision~\cite{picard2021torchmanualseed3407} was conducted.
%
From this study, the relevance and importance of setting a confidence interval to gauge meaningful improvements in the performance of deep neural networks is brought to the forefront.
%
However, in practice it is difficult to define rigorous statistical confidence intervals, as a complete training cycle of both our proposed SOS pre-training and target task takes approximately one week.
Thus, we resort to a pragmatic approach to define a confidence interval.
During the initial stage of this project, we ran our simplest baseline five times with the same hyperparameter configuration, but with different initialization seeds. By contrasting the differences in performance among the multiple performance metrics (KPIs) across multiple runs, we observe that KPIs varied in a range $\pm0.25\%$. Therefore, we deem relevant results those bigger than $0.5\%$.








%
%
\bibliographystyle{splncs04}
\bibliography{egbib}

\end{document}